\documentclass[review]{elsarticle}
\usepackage{graphicx}
\usepackage[ruled,vlined]{algorithm2e}
\usepackage{tabularx}
\usepackage{hyperref}
\usepackage{lineno,hyperref}
\usepackage{lscape}
\modulolinenumbers[5]

\journal{arXiv}

\begin{document}

\begin{frontmatter}

\title{TLA: Twitter Linguistic Analysis}

\author{Tushar Sarkar}
\address{KJ Somaiya College of Engineering, Mumbai }

\author{Nishant Rajadhyaksha}
\address{KJ Somaiya College of Engineering, Mumbai }

\begin{abstract}
Linguistics have been instrumental in developing a deeper understanding of human nature. Words are indispensable to bequeath the thoughts, emotions, and purpose of any human interaction, and critically analyzing these words can elucidate the social and psychological behavior and characteristics of these social animals. Social media has become a platform for human interaction on a large scale and thus gives us scope for collecting and using that data for our study. However, this entire process of collecting, labeling, and analyzing this data iteratively makes the entire procedure cumbersome. To make this entire process easier and structured, we would like to introduce TLA(Twitter Linguistic Analysis). In this paper, we describe TLA and provide a basic understanding of the framework and discuss the process of collecting, labeling, and analyzing data from Twitter for a corpus of languages while providing detailed labeled datasets for all the languages and the models are trained on these datasets. The analysis provided by TLA will also go a long way in understanding the sentiments of different linguistic communities and come up with new and innovative solutions for their problems based on the analysis.
\end{abstract}

\begin{keyword}
TLA, Machine Learning, Analysis, NLP
\end{keyword}

\end{frontmatter}

\section{Introduction}
Language is the fundamental building block upon which communication systems are developed.\cite{downes1998language} Words are necessary to understand the meaning and context of the information being provided over any given subject. Roughly around 6,909 languages are in effect today built around the cultural evolution throughout human history.\cite{anderson2010many} The largest spoken language Mandarin Chinese is spoken by approximately 1.1 billion speakers. Given the amount of information encoded in languages, it is a natural measure to extract data from languages for analysis and to gain a deeper insight into human nature. Twitter is one of the most popular social media applications in use currently.\cite{miller2009s} Its popularity stems from the fact that it provides functionality to its users to broadcast their thoughts within a 280 character limit tweet. Twitter supports multiple languages and has about 199 monetizable users registered.\cite{weerkamp2011people} Hence Twitter can be described as a database containing a great amount of data. This data can be employed for a wide variety of applications. Machine Learning may be applied to learn the semantics and contexts of data hidden behind the words used.\cite{tatar2005word} Machine learning has been successfully applied in topics ranging from Data Mining to  sentiment analysis of different spheres of Twitter.\cite{saif2012semantic} It is also used for Auto-correct applications, developing algorithms to detect vitriolic and abusive messages,fraudulent-user-detection, etc.\cite{wang2017gang}\cite{founta2018large} However, reproducibility of such machine learning techniques is seldom possible for different contexts as topics on Twitter are highly variable. Machine learning Libraries exist for data extraction and analysis but few provide support for multilingual setups. A lightweight library with functionalities aimed to ease usability and for data extraction and analysis of Twitter data can be of great use to harness the power of Twitter data. Implementing such an easy-to-use library can encourage more researchers from a myriad of fields to use textual data for their studies without worrying about Natural Language Processing(NLP). Such a highly scalable library has the potential to improve and attract active contributors and can have massive potential for growth. TLA aims to provide a comprehensive Python library that includes functionalities to extract, process, label, and analyze multilingual datasets.

\section{Methodology}

\subsection{Extracting Data from Twitter}

Twitter contains data in form of tweets. A tweet is a body of characters with an upper limit of 280 characters. A tweet can be composed in multiple languages and is unstructured.\cite{tighe2015painful} 
To build a python library containing datasets, information from the tweets has to be extracted and processed in python.
To achieve this task we used a python library snscrape.\cite{blair2021using} snscrape is a library that contains different functionalities to collect tweets from Twitter. It provides helpful flags to help filter tweets on conditions such as the number of likes on a particular tweet or the number of replies etc. to help filter currently trending tweets on Twitter we used the flags minfaves and has\_engagement. To filter a tweet by the language we used the flag lang. We set a minimum threshold of about 9000 likes using the minfaves flag and set the has\_engagement as "TRUE" to filter out the current trending tweets on Twitter. We selected about 500 trending tweets for each language included in our study by setting the lang flag to the desired language code for eg. lang:'en' to filter English language tweets and iterating through 500 results given as the output by snscrape.

\subsection{Pre-Processing}
Pre-processing was conducted on the data to create a list of processed words the tweet contains. We attempted to remove HTML tags, Unicode characters, symbols, emoticons, punctuations, stopwords, and hyperlinks from tweets to minimize the noise and optimize the semantic and contextual words that help us get a better understanding of the data. We used the python regex module to substitute any occurrence of punctuation marks, HTML tags, and hyperlinks with an empty string. We further used the python library nltk to produce a list of stop words for the languages in our dataset.\cite{loper2002nltk} We removed stop words in our tweet by comparing individual words against the list of stop words. We then created a list of words all in lower case to complete our pre-processing stage..

\subsection{Creating Labeled Data-sets}
To create labeled datasets we filtered tweets by language and stored the tweets as a  file with the .csv file extension for a specific language. We then processed each tweet to understand the sentiment the tweet was trying to express and labeled the respective tweet as Positive if it was trying to express a positive sentiment and Negative if the tweet was trying to express a negative sentiment. We processed about 16 languages and created 16 labeled datasets. The languages are listed in the table below: 

% \usepackage{lscape}
\begin{tabularx}{0.8\textwidth} { 
  | >{\centering\arraybackslash}X 
  | >{\centering\arraybackslash}X 
  | >{\centering\arraybackslash}X 
  | >{\centering\arraybackslash}X | }

\hline 
English        & Hindi          \\ \hline
Chinese        & Urdu           \\ \hline
Thai           & Indonesian     \\ \hline
russian        & Romainan       \\ \hline
Dutch          & Japanese       \\ \hline
French         & Persian        \\ \hline
Portuguese     & Swedish        \\ \hline
 \end{tabularx}
\\

\hspace{25mm} Languages supported by TLA\\ 
\\

\subsection{Language Identification}

We developed a Bert Based architecture to classify languages based on the words contained in the tweet.\cite{devlin2018bert} We then created a file with the .pt extension to save the trained weights from our classifier so that it can be used easily for inference. As Bert is quite a large architecture we also trained a random forest model and saved both the vectorizer and the model as .pkl file.\cite{svetnik2003random}

\subsection{Analysis}
The following analysis was done on the basis of the extracted, processed and labeled tweets: \\

\begin{tabularx}{0.8\textwidth} { 
  | >{\centering\arraybackslash}X 
  | >{\centering\arraybackslash}X 
  | >{\centering\arraybackslash}X 
  | >{\centering\arraybackslash}X | }
 \hline
 Language & Total tweets & Positive Tweets Percentage & Negative Tweets Percentage   \\
 \hline
 English & 500 &  66.8 & 33.2  \\
  \hline
 Spanish & 500 &  61.4 & 38.6 \\
 \hline
 Persian  & 50 &  52 & 48  \\
 \hline
 French & 500 & 53 & 47 \\
 \hline
 Hindi & 500 & 62 & 38 \\
 \hline
 Indonesian & 500 & 63.4 & 36.6 \\
 \hline
 Japanese & 500 & 85.6 &  14.4  \\
 \hline
 Dutch & 500 & 84.2 &  15.8  \\
 \hline
 Portuguese & 500 & 61.2 & 38.8 \\
 \hline
 Romainain & 457 & 85.55 &  14.44 \\
 \hline
 Russian & 213 & 62.91 & 37.08 \\
 \hline
 Swedish & 420 & 80.23 & 19.76 \\
 \hline
 Thai & 424 & 71.46 & 28.53 \\
 \hline
 Turkish & 500 & 67.8 & 32.2  \\
 \hline
 Urdu & 42 & 69.04 &  30.95\\
 \hline
 Chinese & 500 & 80.6 & 19.4  \\
 \hline

 \end{tabularx}
\\

\hspace{33mm} Analysis of tweets \\ 
\\

\section{Impact}
This library makes it very easy for researchers in different fields like psychology, social sciences, etc to use the freely available Twitter data in their studies to analyze the general characteristics of different linguistic communities and make the entire process of inference and analysis easier. It will also help all the computer science researchers to extract and label information in a hassle-free way and also start with the baseline models provided in the library instead of starting from scratch.

\section{Acknowledgements}
I would like to thank Chandan Sarkar, Mallika Sarkar, Amit Rajadhyaksha, Priti Rajadhyaksha, Disha Shah for their constant guidance and valuable feedback. I am also grateful to Aparna Sarkar, Sneha Kothi, Tanvi Rajadhyaksha and the entire community for their priceless suggestions which went a long way for improving the architecture. 

\section*{References}

\bibliography{mybibfile}

\end{document}